\pdfoutput=1
\documentclass[11pt,table,dvipsnames]{article}

\usepackage{acl}

\usepackage{times}
\usepackage{latexsym}

\usepackage[T1]{fontenc}

\usepackage[utf8]{inputenc}

\usepackage{microtype}

\usepackage{inconsolata}

\usepackage{booktabs}
\usepackage{multirow}
\usepackage{amsmath}
\usepackage{amssymb}
\usepackage{subcaption}
\usepackage{enumitem}
\usepackage{graphicx}

\title{Compact Speech Translation Models via Discrete Speech Units Pretraining}

\author{Tsz Kin Lam \and  Alexandra Birch \and  Barry Haddow
  \\School of Informatics, University of Edinburgh \\ \texttt{\normalsize\{tlam, a.birch, bhaddow\}@ed.ac.uk } \\
 }

\begin{document}
\maketitle
\begin{abstract}
We propose a pretraining method to use Self-Supervised Speech (SSS) model to creating more compact Speech-to-text Translation. In contrast to using the SSS model for initialization, our method is more suitable to memory constrained scenario such as on-device deployment. Our method is based on Discrete Speech Units (DSU) extracted from the SSS model. In the first step, our method pretrains two smaller encoder-decoder models on 1) Filterbank-to-DSU (Fbk-to-DSU) and 2) DSU-to-Translation (DSU-to-Trl) data respectively. The DSU thus become the distillation inputs of the smaller models. Subsequently, the encoder from the Fbk-to-DSU model and the decoder from the DSU-to-Trl model are taken to initialise the compact model. Finally, the compact model is finetuned on the paired Fbk-Trl data. In addition to being compact, our method requires no transcripts, making it applicable to low-resource settings. It also avoids speech discretization in inference and is more robust to the DSU tokenization. Evaluation on CoVoST-2 (X-En) shows that our method has consistent improvement over the baseline in three metrics while being compact i.e., only half the SSS model size. 
\end{abstract}

\section{Introduction}\label{sec:intro}
In Speech-to-text Translation (ST), using Self-Supervised Speech (SSS) models, such as wav2vec 2.0 and HuBERT \cite{baevski2020wav2vec,hsu2021hubert}, as model initialization is now common to obtain the SOTA result \cite{agrawal-etal-2023-findings}. 
Nevertheless, such model initialisation makes the ST model less memory-adaptive and could impose a large memory footprint. These factors hinders on-device deployment that is crucial for privacy and useful in the absence of internet connection. 

How can we use the SSS model(s) to create a more compact ST model? When using the SSS model for initialization, the corresponding ST model uses the dense representations of the SSS model for its task. Alternatively, an informative proxy, which requires less memory to obtain, for the dense representation may make the ST model more compact. 

Discrete Speech Units (DSU) extracted from the SSS model can be such a good proxy. DSU are K-Means clusters of speech representations from selected layers of the SSS model. It represents sequence of discrete tokens, which are easier to model within a text processing architecture \cite{polyak21_interspeech,chou-etal-2023-toward}. DSU sequences\footnote{In this paper, DSU and DSU sequences are used interchangeably. When we need to focus on a few units of the sequence, we call them DSU tokens.} are far smaller than the sequences of dense representations. Therefore, a straightforward method to distill the SSS models is to use DSU as speech inputs, aka the DSU-to-Translation (DSU-to-Trl) model. Although using DSU as inputs allows for transfer learning and a memory-adaptive model, using them at inference still requires storing and calling the quantization modules, i.e, the SSS model and the K-Means model. 

We thus propose to use DSU for pretraining (PT) rather than as model input to make ST models more compact. Our method distils the SSS model by pretraining smaller models on the corresponding DSU. More specifically, our method firstly pretrains two smaller encoder-decoder models on 1) Filterbank-to-DSU (Fbk-to-DSU) and 2) DSU-to-Trl data respectively. The DSU thus become the distillation inputs of the smaller models. Subsequently, the encoder from the Fbk-to-DSU model and the decoder from the DSU-to-Trl model are taken to initialise the compact model. Finally, the compact model is finetuned on the paired Fbk-Trl data. Under this formulation, (1) we can use the SSS model to create a ST model that is adaptive to the memory footprint. (2) Our method requires no transcripts, unlike ASR-pretraining, making it applicable to low-resource settings. (3) Our method avoids using the quantization modules in inference. (4) Extensive results also show that our method is more robust to DSU tokenization than the DSU-to-Trl method.  

We evaluate our method on CoVoST-2 \cite{wang21s_interspeech} X-En language directions ($21$ in total) using multilingual ST. By using a HuBERT-Base model to extract the DSU, our method shows strong and consistent improvements in three evaluation metrics with respect to a ST model that is trained from scratch. Our main contributions are: 
\begin{itemize}
    \item We propose a pretraining method to distil the SSS model to creating a more compact ST model. Rather than competing with the SOTA ST models, adaptability to the memory footprint is our key focus. 
    \item Our method uses DSU for pretraining rather than as model inputs. This lowers the inference cost, especially for on-device purpose, by avoiding the quantization modules (storage and running).  
    \item We conduct extensive analysis to study the effect of DSU tokenization to both using DSU as model inputs and as pretraining. Our pretraining method is found to be more robust to different tokenizations.
\end{itemize}

\section{Related Work}
There are a number of related works that use DSU to enhance ST. \citet{fang-feng-2023-back} and \citet{zhang-etal-2023-dub} use DSU to create more training data in a back-translation fashion. \citet{chang2023exploring} and \citet{zhang-etal-2023-dub} explore the replacement of Filterbank by DSU as speech input. Furthermore, \citet{10447926} proposes a multi-tasking learning framework with hard parameter sharing, i.e., using a joint vocabulary for text tokens and DSU, to improve the speech-text modality gap.  In contrast, we use DSU and its translation model for pretraining, resulting in a better Fbk-to-Trl model that has a shorter inference pipeline.

In the case of pretraining, \citet{DBLP:conf/icassp/WuKWHMWA23} use a single Speech-to-DSU model in pretraining for general speech-to-text purposes whereas we tailor the use for ST by using a pair of encoder-decoder models. \citet{zhang-etal-2022-speechut} also decompose ST into speech-to-unit and unit-to-text tasks. Their training is based on masked unit prediction, and it requires an extra unit-encoder module in inference. In contrast, we resort to supervised training on the DSU in acoustic pretraining and require no extra module in inference.  More importantly, our goal is to make (multilingual) ST more compact, aiming  also at low-resource settings where transcripts are not easily available, rather than learning a joint semantic space for both transcripts and audios.

\section{Method}
\begin{figure}[ht]
    \centering
    \includegraphics[width=\linewidth]{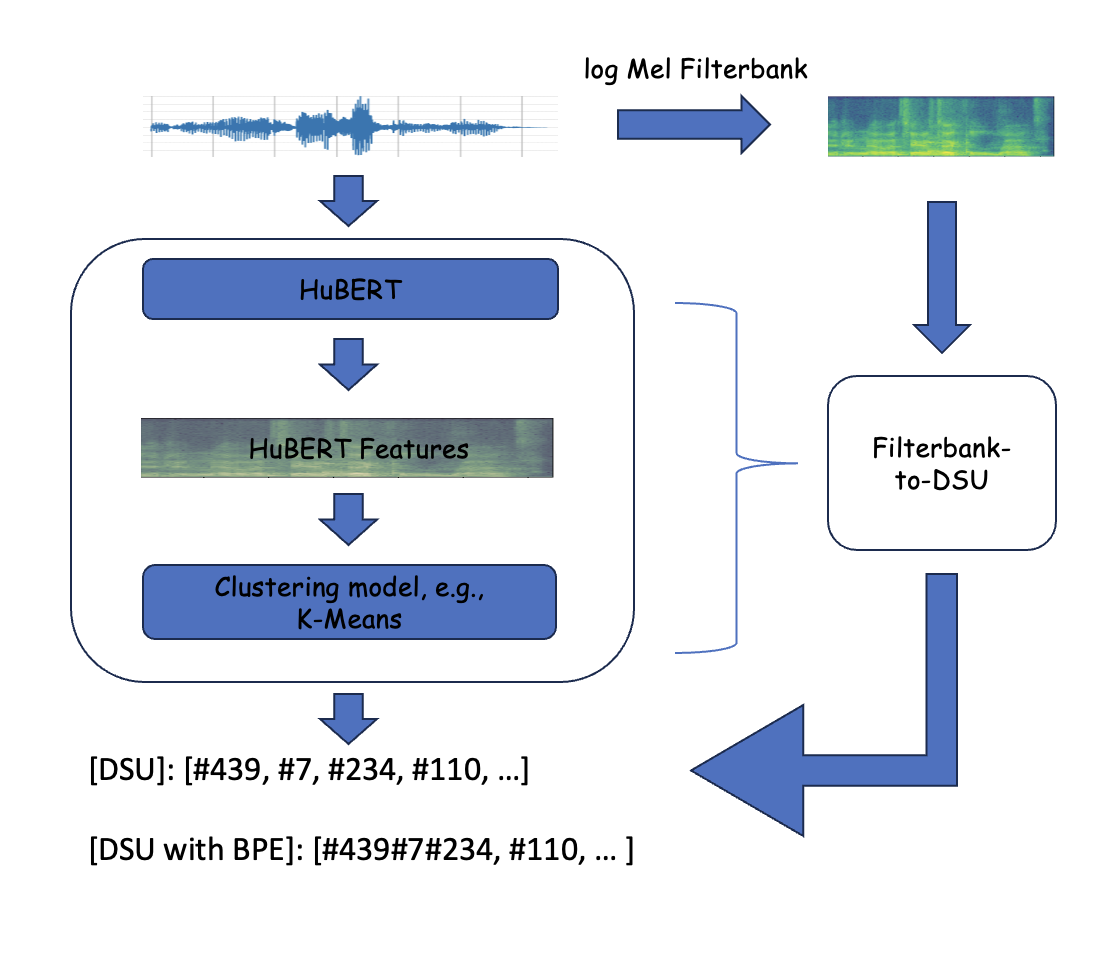}
    \caption{Illustration of the Fbk-to-DSU model. It is like an auto-encoding training process, but between a continuous format (log Mel Filterbank) and its discrete format (DSU) that is extracted from a HuBERT model.}
    \label{fig:fbk_to_dsu}
\end{figure}

Our method uses DSU in the form of pretraining to distil knowledge from the SSS (dense) representations to creating a more compact ST model. 

In the first step, our method pretrains two smaller encoder-decoder models on 1) Fbk-to-DSU and 2) DSU-to-Trl data respectively. The Fbk-to-DSU model takes the log Mel Fbk as the encoder input and predicts the DSU sequence. The model is trained by an interpolation of Connectionist Temporal Classification (CTC, \citet{DBLP:conf/icml/GravesFGS06}) loss that is applied to the last encoder layer and label-smoothed Cross-Entropy (CE) loss:
\begin{equation}
    \mathcal{L}^{\text{Fbk-to-DSU}} = (1-\lambda_{\alpha})\mathcal{L}_{\text{CE}}(\mathbf{U}|\mathbf{F}) + \lambda_{\alpha}\mathcal{L}_{\text{CTC}}(\mathbf{\tilde{U}}|\mathbf{F})
\end{equation}
where $\mathbf{F} \in \mathbb{R}^{T \text{x} D}$, $\mathbf{U} \in \mathcal{U}$ and $\mathbf{\tilde{U}} \in \mathcal{\tilde{U}} = \{\mathcal{U}, \emph{blank}\}$ are the Fbk, DSU and the CTC label sequences respectively. The CTC vocabulary correspond to an union of the same vocabulary used in the CE loss and a $\emph{blank}$ label. The idea is similar to an autoencoder, but the Fbk-to-DSU model is trained to map the Fbk inputs to its discrete form from the SSS model in a multi-task learning fashion (Figure \ref{fig:fbk_to_dsu}). The DSU-to-Trl model learns via CE to predict the translations $\mathbf{Y}$ given $\mathbf{U}$: $\mathcal{L}^{\text{DSU-to-Trl}}= \mathcal{L}_{\text{CE}}(\mathbf{Y}|\mathbf{U})$. In essence, we use the DSU to bridge the speech and text modalities. 

Next, we use the encoder of the Fbk-to-DSU model and the decoder (and its output layer) of the DSU-to-Trl model to initialise the compact model, followed by finetuning on the paired Fbk-Trl data using both CE and CTC loss \cite{gaido-etal-2021-ctc,zhang-etal-2023-efficient} on the translations:
\begin{equation}
\mathcal{L}^{\text{FT}}= (1-\lambda_{\beta})\mathcal{L}_{\text{CE}}(\mathbf{Y}|\mathbf{F}) + \lambda_{\beta}\mathcal{L}_{\text{CTC}}(\mathbf{\tilde{Y}}|\mathbf{F})
\end{equation}
where $\mathbf{\tilde{Y}} \in \mathcal{\tilde{Y}} = \{\mathcal{Y}, \emph{blank}\}$. 

\subsection{Tokenization of DSU in different models} 
\begin{figure}[ht]
    \centering
    \includegraphics[width=\linewidth]{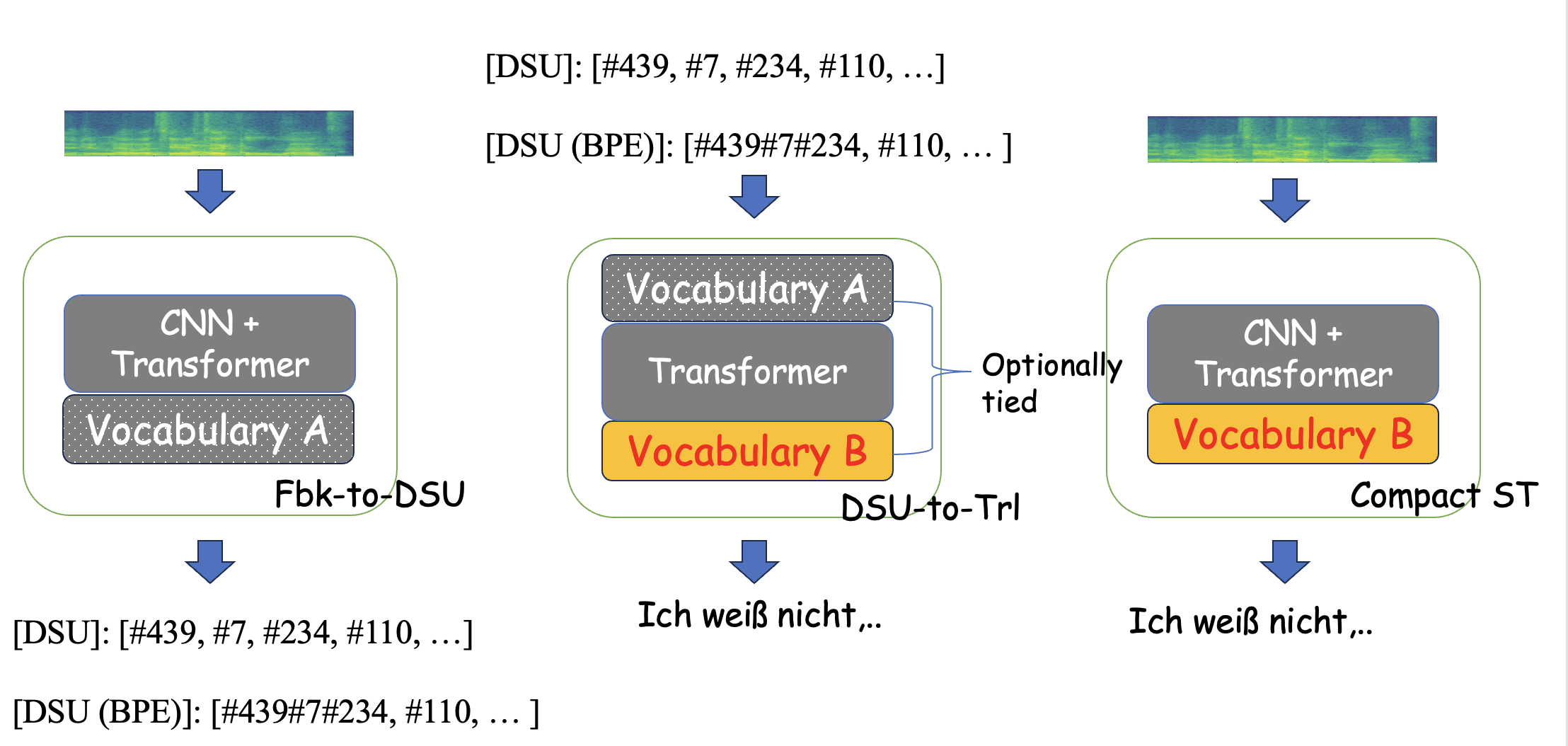}
    \caption{Aligning the DSU tokenization of the Fbk-to-DSU, DSU-to-Trl and compact ST model.}
    \label{fig:dsu_to_trl}
\end{figure}
The discrete nature of DSU makes the above training process similar to the transcripts-based pretraining. However, DSU is self-supervised, whereas transcripts require human annotations. DSU are also much longer and can be represented with various sets of symbols. 

The length issue could be relieved by merging sequential repetitions \cite{ao22_interspeech}, e.g., '\#$1$ \#$1$ \#$1$ \#$456$ \#$456$ \#$23$' becomes '\#$1$ \#$456$ \#$23$', where each DSU token is denoted by a \#\{integer\}. Byte Pair Encoding (BPE) \cite{sennrich-etal-2016-neural} could be applied to reduce the DSU sequence length further, e.g., '\#$1$ \#$456$ \#$23$' could be split into a single subword unit: '\#$1$\#$456$\#$23$'. 

Since both Fbk-to-DSU and DSU-to-Trl models map to different targets, and DSU can be represented with various set of symbols, we align the tokenizations (or called vocabularies\footnote{We use vocabulary and tokenization interchangeably, since we did not apply subword regularisation.}) of the two models. Figure \ref{fig:dsu_to_trl} provides an illustration. The vocabulary of the Fbk-to-DSU model (Vocabulary A) is identical to the source vocabulary of the DSU-to-Trl model (their weights are not shared since these two models are trained independently), whereas the target vocabulary (Vocabulary B) of the DSU-to-Trl model is identical to the target vocabulary of the final compact model (their weights are shared during initialisation). The DSU-to-Trl model is similar to a text translation model, so we also experiment of using separate vocabularies or a joint vocabulary. If a joint vocabulary of English subword units and DSU (BPE or not) is used, all the three models would have the same vocabulary, and the weights of the source and target vocabularies of the DSU-to-Trl model are also tied.

\section{Experiments}
\subsection{Data Preprocessing}
We follow standard practices to preprocess the CoVoST-2 X-En data. For speech inputs using $80$-D log Mel Fbk, we computed the features for every $10$ms with a $25$ms window and then normalized them using its mean and variance computed over each channel. We use the BPE implementation from \textsc{sentencepiece} \cite{kudo-richardson-2018-sentencepiece} and obtain vocabulary of size $8$K on the English target, $16$K on the (non-English) transcripts and $32$K on the DSU, unless otherwise specified.

We use HuBERT-Base\footnote{\url{https://github.com/facebookresearch/fairseq/tree/main/examples/textless_nlp/gslm/speech2unit}} model to extract the DSU by first downsampling the CoVoST-2 audio to $16$KHz. Each audio data utterance is then converted into the DSU, i.e., the clustering indexes, by applying K-Means clustering (K=$1,000$;  \texttt{MiniBatchKMeans} from \textsc{sklearn}) on its HuBERT representation from the $6$th layer \cite{lakhotia-etal-2021-generative}. To train the K-Means model, we divide the $21$ language pairs into three groups: 1) \{ar, cy, et, id, ja, lv, mn, sl, sv, ta, tr\}, 2) \{nl, pt, ru, zh\} and 3) \{ca, de, es, fa, fr, it\}. We then sample $1$K instances for each language pair in group 1), which becomes $3$K in group 2) and $12.5$K in group 3), to create a multilingual training dataset of $98$K instances for the K-Means model. 

\subsubsection{On the choice of using HuBERT-Base}
Given the rapid advance in the SSS models, there are many alternatives, such as XLS-R \cite{babu2021xls} and Wavlm \cite{chen2022wavlm}, for extracting the DSU for our method. These models are larger in scale and could be multilingual, thus providing DSU of higher qualities. The improvement of our method by using DSU from the HuBERT-Base would probably be a lower-bound, considering its relatively poor qualities to the bigger models.
Since our goal is about compactness via DSU pretraining rather than comparing the DSU qualities across the SSS models, we took a simple HuBERT-Base model to illustrate the idea. Pretraining only on English audio data could also suggest hints on whether the DSU and our method could be generalised to languages that are unseen to the SSS models. 

\subsection{Model Configuration}\label{sec:model_config}
All models are based on Transformer \cite{vaswani2017attention} with implementations from \textsc{fairseq} \cite{ott-etal-2019-fairseq,wang-etal-2020-fairseq}. In the Fbk-to-Token (i.e. transcriptions, DSU, or translations) models, the encoder has convolutional layers to downsample the Fbk by a factor of $4$. There are $12$-$6$ layers in the transformer encoder-decoder, whereas the embedding and feed-forward network (FFN) dimensions are $256$ and $4,096$ respectively, unless otherwise specified. It is worth noting that: (1) The Fbk-to-DSU model is not trained on the translations, so it is not directly comparable to the ST models. Its effect on ST lies on its pretrained encoder (Table \ref{tab:tokenization}). (2) The DSU-to-Trl model is a ST model which decoder can be used for initialization. 

\vspace{0.5em}\noindent\textbf{Scratch} is a ST model trained on the paired speech-translation data without pretraining. 

\vspace{0.5em}\noindent\textbf{ASR Pretraining} refers to a ST model whose encoder is initialized by a speech recognition task with CTC regularisation on the transcripts. 

\vspace{0.5em}\noindent\textbf{DSU-to-Trl} follows the Transformer used in text translation. We use $6$-$6$ layers in the encoder-decoder which the dimension of embedding and FFN is $256$ and $2,048$ respectively. In addition, we use "pre" layer-normalization \cite{nguyen-salazar-2019-transformers}. Despite its smaller model size, its inference requires the quantization modules.   
 
\vspace{0.5em}\noindent\textbf{Hu-Transformer} uses the entire HuBERT as the speech encoder initialization \cite{fang-feng-2023-back}. For comparison to our DSU-Adapter, its subsequent encoder-decoder also has $1$-$6$ layers. 

\vspace{0.5em}\noindent\textbf{DSU-Adapter} is our proposed method. To better align the two pre-trained components, we also experiment with adding an extra encoder layer as a simple adapter layer after the pre-trained encoder. Because of the small model size, all model parameters are trainable. Since its decoder is initialized by the DSU-to-Trl method, its decoder FFN dimension is $2,048$.

\vspace{0.5em}\noindent\textbf{Enc-Init} is a ST model that has its encoder initialized by the Fbk-to-DSU encoder. \textbf{EncDec-Init} is a DSU-Adapter model without the adapter layer. 

\subsection{Training and Inference}
It is worth noting that we do not use extra audio data, e.g., Libri-Light \cite{DBLP:conf/icassp/KahnRZKXMKLCFLS20} in our (pretraining) experiments. Furthermore, we apply the following conditions in (pre-)training:  
\begin{itemize}
    \item We skip training data that are longer than $30$ seconds (audio) or $1,024$ target tokens.
    \item We apply SpecAugment \cite{park19e_interspeech} with parameters: \{$F=30, T=40, m_{F}=2, m_{T}=2$\} on Filterbank inputs. 
    \item We share the embedding weights when using a joint vocabulary in the DSU-to-Trl model.
    \item We set $\lambda_{\alpha}$ and $\lambda_{\beta}$ in CTC to $0.3$ and the smoothing parameter to $0.1$
    \item We initialize the encoder (decoder) with the last (best) checkpoint from the PT model.
    \item We use Adam optimizer with inverse square root scheduler for all model training.
    \item In all Fbk-to-Token models, the \emph{effective mini-batch size}, \emph{warm-up steps}, \emph{peak learning rate} and \emph{training steps} are $32$K frames, $25$K, $2$e$-3$ and $60$K steps respectively. 
    \item Similarly, in all DSU-to-Trl models, we use $80$K tokens, $10$K, $5$e$-4$ and $50$K steps. 
    \item Similarly, in Hu-Transformer, we use $4$M frames, $4$K, $1$e$-4$ and $300$K steps. 
\end{itemize}    
In inference, we average the last 5 checkpoints and use beam size of $5$ in generation. All experiments are run on Nvidia A100 GPUs. It takes about $1$ day for $2$ A100 ($40$GB) GPUs to complete an experiment that uses Filterbank as speech inputs. 

\begin{table*}[t]
\centering
\resizebox{\textwidth}{!}
{
\begin{tabular}{l|cccc|cccc|cccc}
\toprule
\multirow{2}{10em}{AST model (\#Params)}  & \multicolumn{4}{c}{BLEU} & \multicolumn{4}{c}{chrF} & \multicolumn{4}{c}{COMET-22-DA} \\
& High & Mid & Low & All & High & Mid & Low & All & High & Mid & Low & All \\
\midrule
Scratch (52M) & $19.4$ & $7.91$ & $0.73$ & $5.99$ & $43.6$ & $27.2$ & $14.6$ & $23.1$ & $0.605$ & $0.498$ & $0.433$ & $0.481$\\
ASR-Pretraining (52M) & $26.5$ & $12.2$ &  $1.82$ & $9.00$ & $51.9$ & $32.8$ & $16.4$ & $27.1$ & $0.680$ & $0.537$ & \underline{$0.443$} & $0.511$\\
Hu-Transformer (113M) & $24.3$ & $11.4$ & \underline{$2.18$} & $8.60$ & $49.9$ & $31.9$ & \underline{$17.0$} & $26.8$ & $0.650$ & $0.522$ & $0.439$ & $0.499$\\
\midrule
DSU-Adapter (48M) & \underline{$26.5$} & \underline{$12.9$} & $1.76$ & \underline{$9.13$} & \underline{$52.1$} & \underline{$33.9$} & $16.5$ & \underline{$27.4$} & \underline{$0.681$} & \underline{$0.548$} & $0.442$ & \underline{$0.513$} \\
\bottomrule
\end{tabular}
}
\caption{Results in BLEU, chrF and COMET-22-DA on the test set of CoVoST-2 (X-En) by resource group. In all metrics, DSU-Adapter is much better than Hu-Transformer, which is $2.3$ times larger, in both "High" and "Mid" groups. DSU-Adapter, which does not requires transcripts in training, is also on a par with ASR-Pretraining. The best result in each group is denoted by '\_'. 
}
\label{tab:covost2}
\end{table*}

\section{Results and Analysis}\label{sec:results_and_analysis}
Before discussing the results, it is worth noting that (1) \emph{Hu-Transformer is not memory-adaptive}, and (2) \emph{ASR-Pretraining requires transcripts, unlike DSU which is self-supervised}. Both methods are introduced for reference purposes of if such resources are available. 

\subsection{Improvement brought by DSU-Adapter}
We divide the 21 language pairs by resource level into: 1) "High": \{ca, de, es, fr\}, 2) "Mid": \{fa, it, pt, ru and zh\}, 3) "Low": \{ar, cy, et, id, ja, lv, mn, nl, sl, sv, ta, tr\} and 4) "All": the 21 languages pairs. We report the average BLEU\footnote{nrefs:1|case:mixed|eff:no|tok:13a|smooth:exp|version:2.3.1} and chrF\footnote{nrefs:1|case:mixed|eff:yes|nc:6|nw:0|space:no|version:2.3.1} over the test sets of each group using \textsc{sacrebleu} \cite{post-2018-call}. In addition, we also provide the result in WMT22-COMET-DA \cite{rei-etal-2022-comet}, which the source inputs are the gold-reference transcripts. 

Table \ref{tab:covost2} compares our DSU-Adapter and the baselines. Our DSU-Adapter is $3$ BLEU (in the group "All") higher than the Scratch model. This shows that our proposed method of using DSU-pretraining can strengthen direct end-to-end ST without requiring transcripts and remain flexible in memory footprint (smaller in size than the HuBERT model). Furthermore, it is better than Hu-Transformer in spite of having half the parameters. For "Mid" and "High", the improvement in BLEU is $1.49$ and $2.23$ points respectively, but it falls short by $0.42$ points for "Low". We also compare to ASR pretraining, which is not always applicable, e.g., in low-resource setting or perhaps even in an unwritten language \cite{zhang2022revisiting}. 
Surprisingly, our adapter is on a par with it, and its BLEU is $0.13$ points better. The result remains consistent when it is measured in chrF and COMET.

\subsubsection{Language-specific performance} 
\begin{figure*}[ht]
    \centering
    \begin{subfigure}{0.75\linewidth}
        \includegraphics[width=\linewidth]{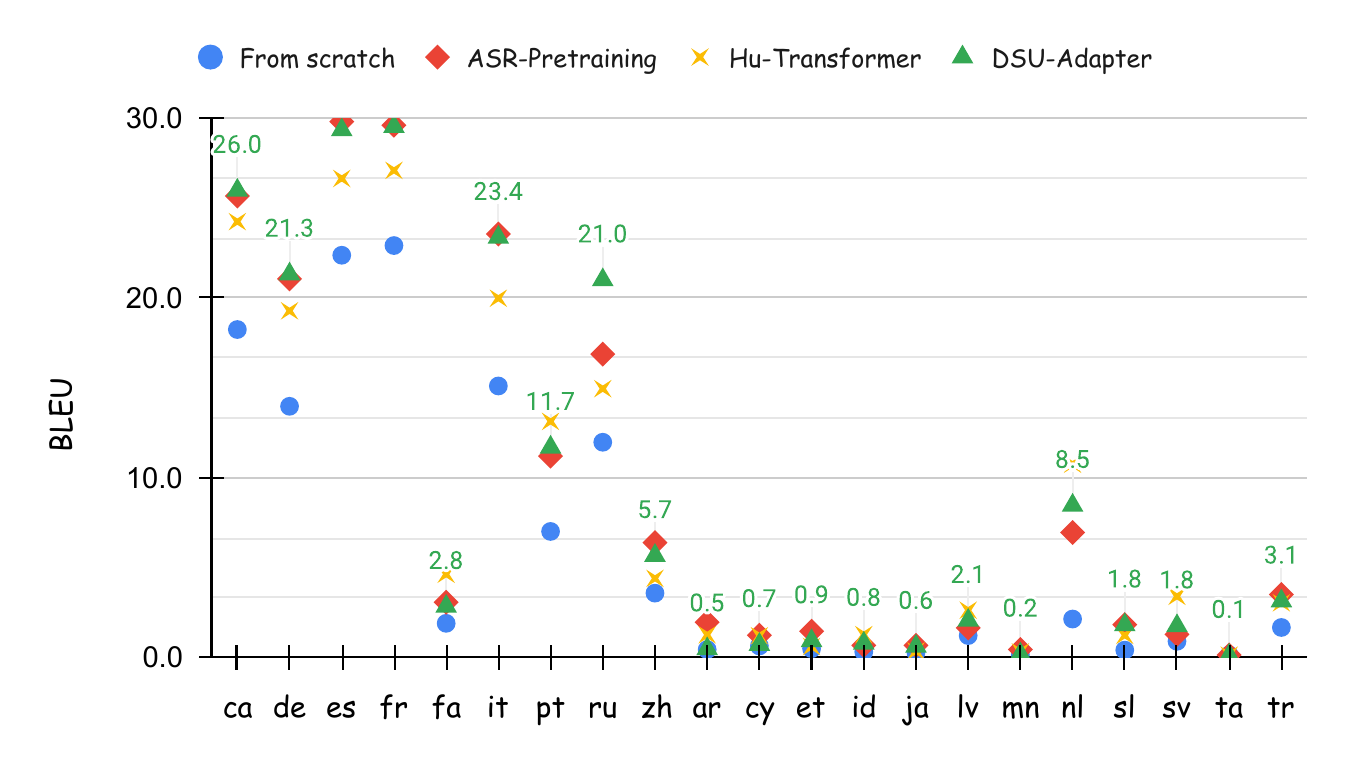}
        \label{fig:bleu_by_lang}
    \end{subfigure}
    \begin{subfigure}{0.75\linewidth}
        \includegraphics[width=\linewidth]{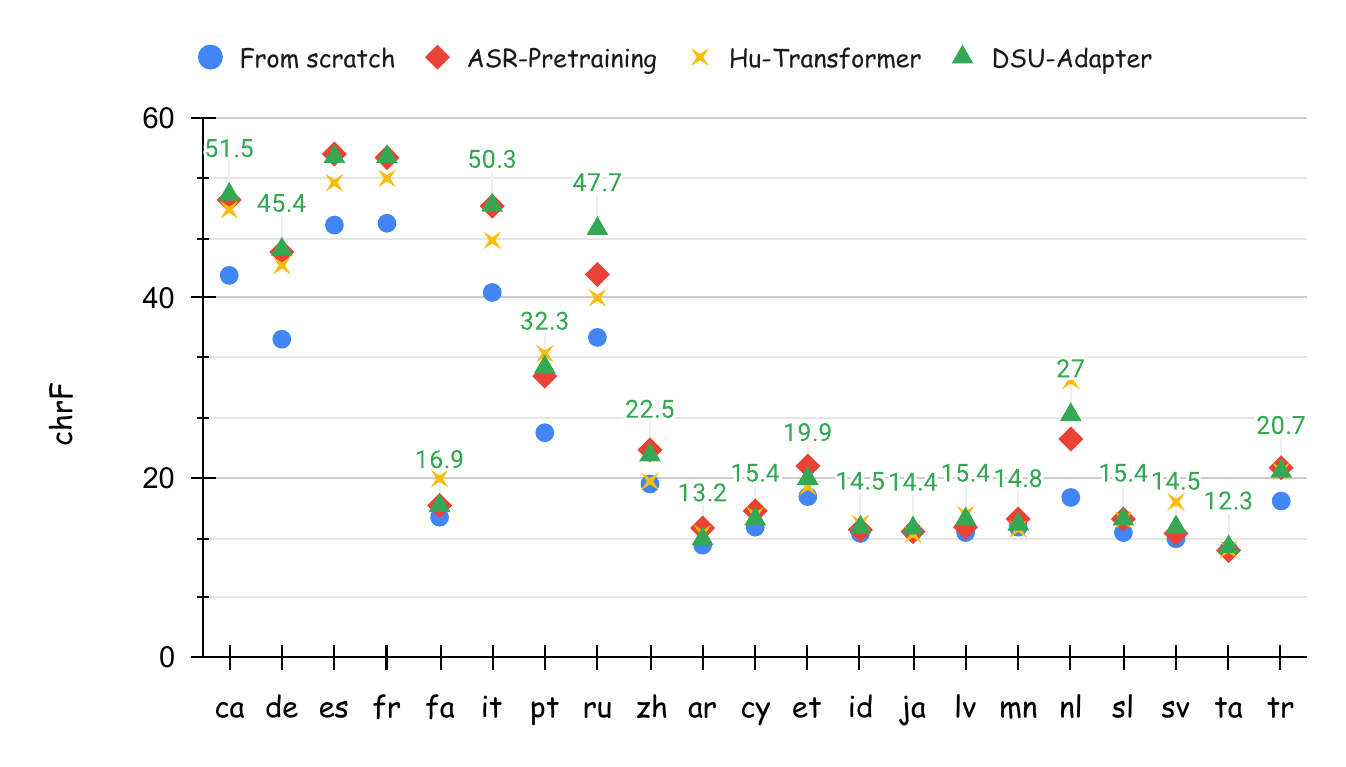}
        \label{fig:chrf_by_lang}
    \end{subfigure}
    \begin{subfigure}{0.75\linewidth}
        \includegraphics[width=\linewidth]{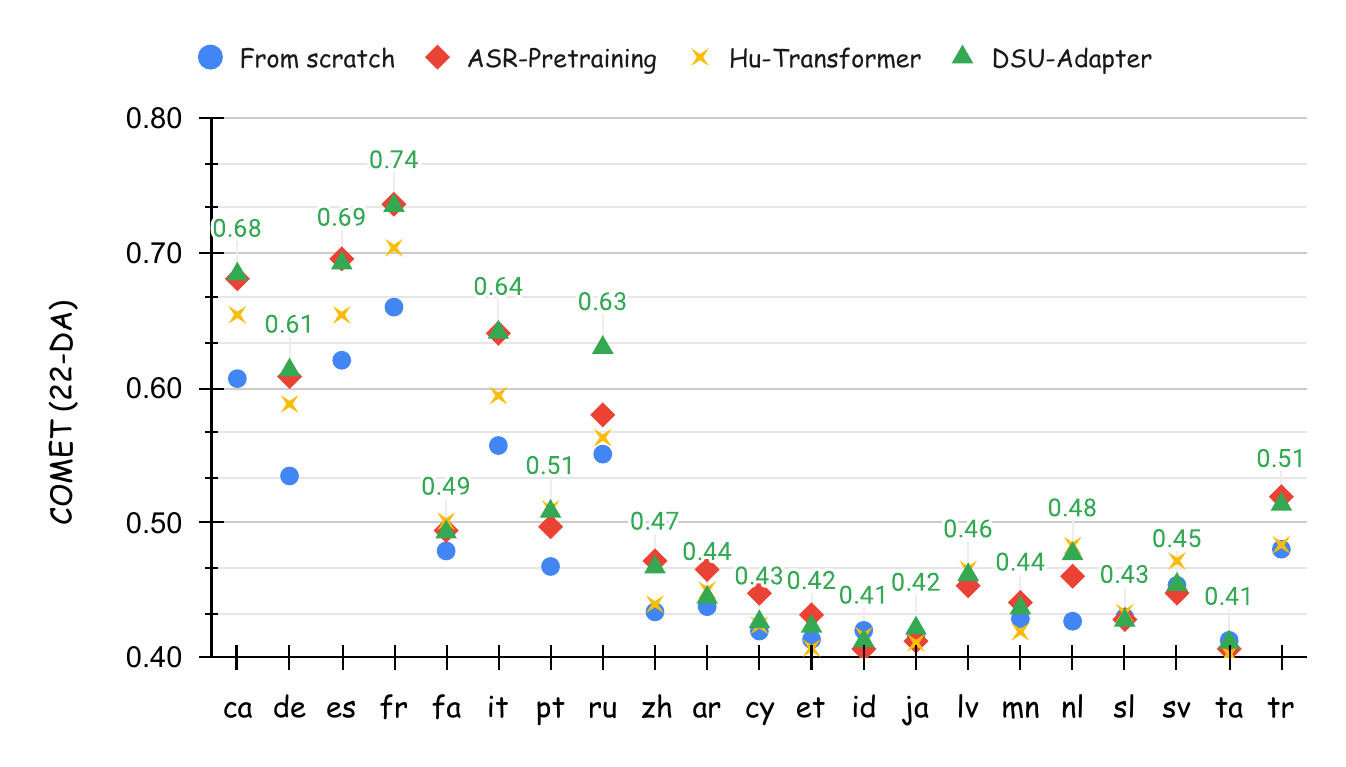}
        \label{fig:chrf_by_lang}
    \end{subfigure}
    \caption{Results in BLEU, chrF and COMET-22-DA on each language pair of CoVoST-2 (X-En).}
    \label{fig:language_specific}
\end{figure*}

Figure \ref{fig:language_specific} shows the performance on each language pair in BLEU, chrF and COMET-22-DA. Our DSU-Adapter (in green triangles) show consistent improvement over the Scratch model (in blue circles) in all language pairs. Such improvement is rather surprising since HuBERT-Base was trained solely on English audio data. We hypothesized that the cross-lingual improvement is related to HuBERT's ability to capture language independent features, e.g. phonetic properties \cite{pasad2023comparative}.

Compared with Hu-Transformer, DSU-Adapter maintains an evident improvement over most language pairs in both "High" and "Mid" groups. Exceptions are in "fa" and "pt", but the lags are almost negligible. In group "Low", Hu-Transformer is slightly better, especially in "nl" and "sv" pairs. However, most translation in this group is barely around $2$ BLEU, and the lags are small.

In most language pairs, DSU-Adapter performs similarly to ASR-pretraining (in red diamonds), except translating from "ru" audios. The improvement in this "ru-en" pair makes DSU-Adapter to have an evident advantage of $0.7$ BLEU in the group "Mid".

\begin{table*}[th]
\centering
\resizebox{\textwidth}{!}
{
\begin{tabular}{c|c|cc|c|cc|c}
\toprule
Has BPE & $|\mathcal{V}|$ & DSU & Length & DSU-to-Trl & Enc-Init & EncDec-Init & DSU-Adapter \\
on DSU? &  & Length & Ratio & ($20$M to $27$M) & ($52$M to $70$M) & ($46$M to $64$M) & ($48$M to $67$M) \\
\midrule
\multirow{6}{1em}{No} & $1$K-$8$K & $176$ & $12.9$ & \cellcolor{LimeGreen} $6.73$ & \cellcolor{LimeGreen} $7.70$ & \cellcolor{LimeGreen} $7.87$ & \cellcolor{SpringGreen} $8.54$\\ 
& $1$K-$16$K & " & $14.1$ & \cellcolor{LimeGreen} $6.36$ & \cellcolor{LimeGreen} $7.50$ & \cellcolor{SpringGreen} $8.00$ & \cellcolor{SpringGreen} $8.43$ \\
& $1$K-$32$K & " & $14.9$ & \cellcolor{LimeGreen} $6.30$ & \cellcolor{LimeGreen} $7.23$ & \cellcolor{LimeGreen} $7.65$ & \cellcolor{LimeGreen} $7.94$  \\ 
\cmidrule(lr){2-8}
& $8$K & " & $12.7$ & \cellcolor{LimeGreen} $6.88$ & \cellcolor{LimeGreen} $7.64$ & \cellcolor{SpringGreen} $8.05$ & \cellcolor{SpringGreen} $8.26$  \\ 
& $16$K & " & $14.0$ & \cellcolor{LimeGreen} $6.33$ & \cellcolor{LimeGreen} $7.41$ & \cellcolor{LimeGreen} $7.91$ & \cellcolor{SpringGreen} $8.17$ \\ 
& $32$K & " & $14.9$ & \cellcolor{LimeGreen} $6.26$ & \cellcolor{LimeGreen} $6.66$ & \cellcolor{LimeGreen} $7.41$ & \cellcolor{LimeGreen} $7.68$ \\ 
\midrule
\multirow{6}{1em}{Yes} & $1$K-$8$K & $221$ & $16.3$ & \cellcolor{OliveGreen} $4.52$ & \cellcolor{SpringGreen} $8.23$ & \cellcolor{SpringGreen} $8.44$ & \cellcolor{SpringGreen} $8.61$ \\ 
& $16$K-$8$K & $129$ & $9.5$ & \cellcolor{OliveGreen} $5.06$ & \cellcolor{SpringGreen} $8.51$ & \cellcolor{SpringGreen} $8.76$ & \cellcolor{SpringGreen} $8.95$ \\ 
& $32$K-$8$K & $115$ & $8.5$ & \cellcolor{OliveGreen} $4.43$ & \cellcolor{SpringGreen} $8.67$ & \cellcolor{yellow} $9.02$ & \cellcolor{yellow} $9.13$ \\ 
\cmidrule(lr){2-8}
& $8$K & $150$ & $7.6$ & \cellcolor{LimeGreen} $7.02$ & \cellcolor{SpringGreen} $8.33$ & \cellcolor{SpringGreen} $8.51$ & \cellcolor{SpringGreen} $8.82$ \\ 
& $16$K & $133$ & $7.7$ & \cellcolor{LimeGreen} $6.50$ & \cellcolor{SpringGreen} $8.57$ & \cellcolor{SpringGreen} $8.61$ & \cellcolor{SpringGreen} $8.93$ \\ 
& $32$K & $118$ & $7.8$ & \cellcolor{OliveGreen} $5.07$ & \cellcolor{SpringGreen} $8.30$ & \cellcolor{SpringGreen} $8.44$ & \cellcolor{SpringGreen} $8.70$ \\ 
\bottomrule
\end{tabular}
}
\caption{(DSU) tokenization effect on 4 ST methods. Each ST model's performance on the CoVoST-2 test set is measured by BLEU on group "All". All 4 methods could perform better than the Scratch model of $5.99$ BLEU as shown on Table \ref{tab:covost2}. In general, darker (brighter) cells refer to weaker (stronger) models. The best two models apply both BPE on the DSU and separate vocabularies in PT (cells in yellow).}
\label{tab:tokenization}
\end{table*}

\subsection{Tokenization effect to the DSU-to-Trl method and the DSU-Adapter method}
In this section, we investigate how tokenization, including BPE, affects the DSU-to-Trl method and the DSU-Adapter method. We are particularly interested in their robustness toward the tokenization, especially using BPE on the DSU, since tuning the quantization process and retraining the subsequent models is computationally expensive. 

In Table \ref{tab:tokenization}, the 1st column "Has BPE on DSU?" indicates if BPE is applied on the DSU. If "Yes", multiple DSU could be merged into one subword unit, e.g., '\#1 \#456 \#23 \#999' could be merged into '\#1\#456\#23\#999'. The 2nd column "$|\mathcal{V}|$" shows the vocabulary configuration: its size, and if the model has a joint vocabulary. For example, "$1$K-$8$K" means that we use a vocabulary of size $1$K for DSU and a second vocabulary of size $8$K for English so that the DSU-to-Trl model would have separate vocabularies for the source (DSU) and target (English) sides. All results are in BLEU averaged over all language pairs, i.e., group "All".

\subsubsection{DSU-to-Trl: robust to tokenization?}
When BPE is not applied on the DSU, those 6 DSU-to-Trl models have $6.48$ $\pm$ $0.26$ BLEU. Despite having smaller model size (<$30$M), they are better than the Scratch model of $5.99$ BLEU. 

When BPE is applied, the sequence length of DSU (DSU Length) could be shortened, which could in turn improve the performance, e.g. the best DSU-to-Trl model happens at configuration "$8$K" with $7.02$ BLEU. However, the DSU-to-Trl method is quite unstable to the use of BPE, as reflected by the $5.12$ $\pm$ $0.83$ BLEU in the other 5 configurations. The correlation between the DSU sequence length, the source-target length ratio, and the ST performance is also not straightforward. For an example, the "$32$K" model (DSU length of $118$) is about $2$ BLEU behind to the "$8$K" model (DSU length of $150$). Therefore, applying BPE on the DSU for length reduction should remain cautious.

\subsubsection{The DSU-Adapter is more robust} 
Unlike DSU-to-Trl method, DSU-Adapter benefits more when BPE is applied to the DSU. Our proposed method has $8.86$ $\pm$ $0.19$ BLEU (over the 6 corresponding configurations), as opposed to $8.17$ $\pm$ $0.32$ BLEU when BPE is not applied. This observation is opposed to the DSU-to-Trl method which only scores $5.54$ $\pm$ $1.07$ BLEU (with also larger variance) when BPE is applied on the DSU but $6.48$ $\pm$ $0.26$ when BPE is not used. The improved mean score and its smaller variance suggests that the DSU-Adapter method is more (DSU) tokenization robust. We see this as a benefit of introducing the DSU, i.e., the SSS model knowledge, via PT rather than as model inputs. 

On top of applying BPE on the DSU, using separate vocabularies in PT is preferred (the two yellow cells on Table \ref{tab:tokenization}) since it performs slightly better, and the DSU, which are not needed in the ST output, would not occupy the target vocabulary.

\subsubsection{Ablation: initialisation in DSU-Adapter}
Having similar model sizes, e.g. about $50$M parameters (Table \ref{tab:tokenization}), DSU-Adapter is better than both EncDec-Init and Enc-Init methods. The translation performance in BLEU (averaged over the 12 vocabularies) is $8.51$ $\pm$ $0.44$, $8.22$ $\pm$ $0.51$, and $7.99$ $\pm$ $0.61$ respectively. Encoder-initialization seems more crucial than decoder-initialization, as reflected by the fact that the best DSU-Adapter model comes from a combination with the weakest DSU-to-Trl model of $4.43$ BLEU.

\subsection{Is CTC applicable also to DSU?}
\begin{table}[h]
\centering
\footnotesize
{
\begin{tabular}{cc|cccc}
\toprule
\multicolumn{2}{c}{Has CTC in} & \multirow{2}{1em}{High} & \multirow{2}{1em}{Mid} & \multirow{2}{1em}{Low} & \multirow{2}{1em}{All} \\
DSU PT? & ST FT? & & & & \\
\midrule
No & No & $25.81$ & $9.91$ & $1.46$ & $8.14$ \\
No & Yes & $26.10$ & $11.35$ & $1.69$ & $8.71$ \\
Yes & No & $25.94$ & $10.82$ & $1.51$ & $8.44$ \\
Yes & Yes & \underline{$26.12$} & \underline{$11.53$} & \underline{$1.73$} & \underline{$8.74$} \\
\bottomrule
\end{tabular}
}
\caption{Effect of CTC on Fbk-to-DSU PT and/or ST FT to the DSU-Adapter method. All results are in BLEU and the best in each group is denoted by '\_'.}
\label{tab:CTC}
\end{table}

Similar to ST methods that use pretrained components, our method could be limited by the \emph{pretraining modality gap} \cite{liu2020bridging,pmlr-v202-le23a}. Motivated by prior works, we investigate mitigating it with CTC. A crucial difference to the prior works is that our method uses DSU for pre-training rather than transcripts. 

We thus study applying CTC in our method at different training stages. Owing to the large number of vocabulary configurations on Table \ref{tab:tokenization}, we only experiment with: 1) "No-BPE $1$K-$8$K", 2) "BPE $8$K", 3) "BPE $32$K" and 4) "BPE $32$K-$8$K". In each training stage, we report the effect of CTC to the ST performance (per resource group) by averaging the BLEU of these $4$ configurations.

Table \ref{tab:CTC} presents the analysis of applying CTC on our DSU-Adapter method. The training condition "Has CTC in DSU PT" refers to the case of applying CTC on the \emph{discrete speech units} in Fbk-to-DSU pretraining, whereas "Has CTC in ST FT" refers to the case of applying CTC on the \emph{translations} in ST finetuning, i.e., on the paired Fbk-Trl data. Our result shows that CTC helps on either stage,  but the gain is $0.27$ BLEU more in ST finetuning. Using them jointly still helps, but the marginal gain is barely $0.03$ BLEU.  


\section{Limitations and future works}


In the previous sections,  we discuss the noticeable benefits of our DSU-pretraining method in creating a more compact ST model. In spite of this, there are several factors that are not thoroughly explored and could improve the model performance further:

\paragraph{K-Means clustering} We did not inspect the clustering size (fixed to $1,000$) and the number of training instances (only fixed to $98,000$) used in training the K-Means clustering model. Apart from tuning its hyper-parameters, using other techniques, such as residual vector quantisation \cite{zeghidour2021soundstream,defossez2022high} and multiple codebooks \cite{guo2023predicting}, might bring better improvement. 

\paragraph{Other acoustic encoders} We did not experiment other acoustic encoders, such as conformer \cite{gulati20_interspeech,papi2023good} and E-Branchformer \cite{peng23b_interspeech}. This stronger encoders should provide further gains for our method since they also enjoy the benefit of pretraining.

\paragraph{A stronger pretrained decoder} Apart from strengthening the encoder, the DSU-to-Trl model and hence its decoder (used in initialisation) could also be improved, e.g. via back-translation, up-sampling the textual sequence \cite{10447926} and pretraining with more text data, while maintaining the small decoder size. 

\paragraph{Further analyses} In addition to improving our pretraining method for better model compactness, there are other related research directions worth further analyzing. One direction would be how, in terms of acoustic pretraining, DSU compared with transcripts (if available in that language) over different data scales. Another interesting research direction would be the comparison and analysis of using DSU or dense features in a large pretrained model setting, such as Whisper \cite{radford2023robust} and Large Language Models. 

\section{Conclusion}
In this paper, we consider a memory-constrained setting for ST. Our proposed method uses DSU in the form of pretraining to distil the knowledge from the Self-Supervised Speech model to creating more compact Speech-to-text Translation. Our compact model, i.e., the DSU-Adapter, shows strong and consistent improvements in three evaluation metrics over the baselines. In contrast to using DSU as model inputs, our method does not require quantization modules in inference and shows stronger robustness to the DSU tokenization. Finally, our method requires no transcripts, making it also suitable for low-resource setting. 

\section*{Acknowledgements}
This work was funded by UK Research and Innovation (UKRI) under the UK
government’s Horizon Europe funding guarantee (grant number 10039436: UTTER). The computations described in this research were performed using the Baskerville Tier 2 HPC service (https://www.baskerville.ac.uk/). Baskerville was funded by the EPSRC and UKRI through the World Class Labs scheme (EP/T022221/1) and the Digital Research Infrastructure programme (EP/W032244/1) and is operated by Advanced Research Computing at the University of Birmingham.

\clearpage


\bibliography{anthology,custom}

\begin{thebibliography}{40}
\expandafter\ifx\csname natexlab\endcsname\relax\def\natexlab#1{#1}\fi

\bibitem[{Agarwal et~al.(2023)Agarwal, Agrawal, Anastasopoulos, Bentivogli, Bojar, Borg, Carpuat, Cattoni, Cettolo, Chen, Chen, Choukri, Chronopoulou, Currey, Declerck, Dong, Duh, Est{\`e}ve, Federico, Gahbiche, Haddow, Hsu, Mon~Htut, Inaguma, Javorsk{\'y}, Judge, Kano, Ko, Kumar, Li, Ma, Mathur, Matusov, McNamee, P.~McCrae, Murray, Nadejde, Nakamura, Negri, Nguyen, Niehues, Niu, Kr.~Ojha, E.~Ortega, Pal, Pino, van~der Plas, Pol{\'a}k, Rippeth, Salesky, Shi, Sperber, St{\"u}ker, Sudoh, Tang, Thompson, Tran, Turchi, Waibel, Wang, Watanabe, and Zevallos}]{agrawal-etal-2023-findings}
Milind Agarwal, Sweta Agrawal, Antonios Anastasopoulos, Luisa Bentivogli, Ond{\v{r}}ej Bojar, Claudia Borg, Marine Carpuat, Roldano Cattoni, Mauro Cettolo, Mingda Chen, William Chen, Khalid Choukri, Alexandra Chronopoulou, Anna Currey, Thierry Declerck, Qianqian Dong, Kevin Duh, Yannick Est{\`e}ve, Marcello Federico, Souhir Gahbiche, Barry Haddow, Benjamin Hsu, Phu Mon~Htut, Hirofumi Inaguma, D{\'a}vid Javorsk{\'y}, John Judge, Yasumasa Kano, Tom Ko, Rishu Kumar, Pengwei Li, Xutai Ma, Prashant Mathur, Evgeny Matusov, Paul McNamee, John P.~McCrae, Kenton Murray, Maria Nadejde, Satoshi Nakamura, Matteo Negri, Ha~Nguyen, Jan Niehues, Xing Niu, Atul Kr.~Ojha, John E.~Ortega, Proyag Pal, Juan Pino, Lonneke van~der Plas, Peter Pol{\'a}k, Elijah Rippeth, Elizabeth Salesky, Jiatong Shi, Matthias Sperber, Sebastian St{\"u}ker, Katsuhito Sudoh, Yun Tang, Brian Thompson, Kevin Tran, Marco Turchi, Alex Waibel, Mingxuan Wang, Shinji Watanabe, and Rodolfo Zevallos. 2023.
\newblock \href {https://doi.org/10.18653/v1/2023.iwslt-1.1} {{FINDINGS} {OF} {THE} {IWSLT} 2023 {EVALUATION} {CAMPAIGN}}.
\newblock In \emph{Proceedings of the 20th International Conference on Spoken Language Translation (IWSLT 2023)}, pages 1--61, Toronto, Canada (in-person and online). Association for Computational Linguistics.

\bibitem[{Ao et~al.(2022)Ao, Zhang, Zhou, Liu, Li, Ko, Dai, Li, Qian, and Wei}]{ao22_interspeech}
Junyi Ao, Ziqiang Zhang, Long Zhou, Shujie Liu, Haizhou Li, Tom Ko, Lirong Dai, Jinyu Li, Yao Qian, and Furu Wei. 2022.
\newblock \href {https://doi.org/10.21437/Interspeech.2022-10368} {{Pre-Training Transformer Decoder for End-to-End ASR Model with Unpaired Speech Data}}.
\newblock In \emph{Proc. Interspeech 2022}, pages 2658--2662.

\bibitem[{Babu et~al.(2021)Babu, Wang, Tjandra, Lakhotia, Xu, Goyal, Singh, von Platen, Saraf, Pino et~al.}]{babu2021xls}
Arun Babu, Changhan Wang, Andros Tjandra, Kushal Lakhotia, Qiantong Xu, Naman Goyal, Kritika Singh, Patrick von Platen, Yatharth Saraf, Juan Pino, et~al. 2021.
\newblock Xls-r: Self-supervised cross-lingual speech representation learning at scale.
\newblock \emph{arXiv preprint arXiv:2111.09296}.

\bibitem[{Baevski et~al.(2020)Baevski, Zhou, Mohamed, and Auli}]{baevski2020wav2vec}
Alexei Baevski, Yuhao Zhou, Abdelrahman Mohamed, and Michael Auli. 2020.
\newblock wav2vec 2.0: A framework for self-supervised learning of speech representations.
\newblock \emph{Advances in neural information processing systems}, 33:12449--12460.

\bibitem[{Chang et~al.(2023)Chang, Yan, Choi, Jung, Lu, Maiti, Sharma, Shi, Tian, Watanabe et~al.}]{chang2023exploring}
Xuankai Chang, Brian Yan, Kwanghee Choi, Jeeweon Jung, Yichen Lu, Soumi Maiti, Roshan Sharma, Jiatong Shi, Jinchuan Tian, Shinji Watanabe, et~al. 2023.
\newblock Exploring speech recognition, translation, and understanding with discrete speech units: A comparative study.
\newblock \emph{arXiv preprint arXiv:2309.15800}.

\bibitem[{Chen et~al.(2022)Chen, Wang, Chen, Wu, Liu, Chen, Li, Kanda, Yoshioka, Xiao et~al.}]{chen2022wavlm}
Sanyuan Chen, Chengyi Wang, Zhengyang Chen, Yu~Wu, Shujie Liu, Zhuo Chen, Jinyu Li, Naoyuki Kanda, Takuya Yoshioka, Xiong Xiao, et~al. 2022.
\newblock Wavlm: Large-scale self-supervised pre-training for full stack speech processing.
\newblock \emph{IEEE Journal of Selected Topics in Signal Processing}, 16(6):1505--1518.

\bibitem[{Chou et~al.(2023)Chou, Chien, Hsu, Livescu, Babu, Conneau, Baevski, and Auli}]{chou-etal-2023-toward}
Ju-Chieh Chou, Chung-Ming Chien, Wei-Ning Hsu, Karen Livescu, Arun Babu, Alexis Conneau, Alexei Baevski, and Michael Auli. 2023.
\newblock \href {https://doi.org/10.18653/v1/2023.findings-emnlp.438} {Toward joint language modeling for speech units and text}.
\newblock In \emph{Findings of the Association for Computational Linguistics: EMNLP 2023}, pages 6582--6593, Singapore. Association for Computational Linguistics.

\bibitem[{D{\'e}fossez et~al.(2022)D{\'e}fossez, Copet, Synnaeve, and Adi}]{defossez2022high}
Alexandre D{\'e}fossez, Jade Copet, Gabriel Synnaeve, and Yossi Adi. 2022.
\newblock High fidelity neural audio compression.
\newblock \emph{arXiv preprint arXiv:2210.13438}.

\bibitem[{Fang and Feng(2023)}]{fang-feng-2023-back}
Qingkai Fang and Yang Feng. 2023.
\newblock \href {https://doi.org/10.18653/v1/2023.acl-long.251} {Back translation for speech-to-text translation without transcripts}.
\newblock In \emph{Proceedings of the 61st Annual Meeting of the Association for Computational Linguistics (Volume 1: Long Papers)}, pages 4567--4587, Toronto, Canada. Association for Computational Linguistics.

\bibitem[{Gaido et~al.(2021)Gaido, Cettolo, Negri, and Turchi}]{gaido-etal-2021-ctc}
Marco Gaido, Mauro Cettolo, Matteo Negri, and Marco Turchi. 2021.
\newblock \href {https://doi.org/10.18653/v1/2021.eacl-main.57} {{CTC}-based compression for direct speech translation}.
\newblock In \emph{Proceedings of the 16th Conference of the European Chapter of the Association for Computational Linguistics: Main Volume}, pages 690--696, Online. Association for Computational Linguistics.

\bibitem[{Graves et~al.(2006)Graves, Fern{\'{a}}ndez, Gomez, and Schmidhuber}]{DBLP:conf/icml/GravesFGS06}
Alex Graves, Santiago Fern{\'{a}}ndez, Faustino~J. Gomez, and J{\"{u}}rgen Schmidhuber. 2006.
\newblock \href {https://doi.org/10.1145/1143844.1143891} {Connectionist temporal classification: labelling unsegmented sequence data with recurrent neural networks}.
\newblock In \emph{Machine Learning, Proceedings of the Twenty-Third International Conference {(ICML} 2006), Pittsburgh, Pennsylvania, USA, June 25-29, 2006}, volume 148 of \emph{{ACM} International Conference Proceeding Series}, pages 369--376. {ACM}.

\bibitem[{Gulati et~al.(2020)Gulati, Qin, Chiu, Parmar, Zhang, Yu, Han, Wang, Zhang, Wu, and Pang}]{gulati20_interspeech}
Anmol Gulati, James Qin, Chung-Cheng Chiu, Niki Parmar, Yu~Zhang, Jiahui Yu, Wei Han, Shibo Wang, Zhengdong Zhang, Yonghui Wu, and Ruoming Pang. 2020.
\newblock \href {https://doi.org/10.21437/Interspeech.2020-3015} {{Conformer: Convolution-augmented Transformer for Speech Recognition}}.
\newblock In \emph{Proc. Interspeech 2020}, pages 5036--5040.

\bibitem[{Guo et~al.(2023)Guo, Yang, Wang, Kong, Yao, Cui, Kuang, Kang, Lin, Luo et~al.}]{guo2023predicting}
Liyong Guo, Xiaoyu Yang, Quandong Wang, Yuxiang Kong, Zengwei Yao, Fan Cui, Fangjun Kuang, Wei Kang, Long Lin, Mingshuang Luo, et~al. 2023.
\newblock Predicting multi-codebook vector quantization indexes for knowledge distillation.
\newblock In \emph{ICASSP 2023-2023 IEEE International Conference on Acoustics, Speech and Signal Processing (ICASSP)}, pages 1--5. IEEE.

\bibitem[{Hsu et~al.(2021)Hsu, Bolte, Tsai, Lakhotia, Salakhutdinov, and Mohamed}]{hsu2021hubert}
Wei-Ning Hsu, Benjamin Bolte, Yao-Hung~Hubert Tsai, Kushal Lakhotia, Ruslan Salakhutdinov, and Abdelrahman Mohamed. 2021.
\newblock Hubert: Self-supervised speech representation learning by masked prediction of hidden units.
\newblock \emph{IEEE/ACM Transactions on Audio, Speech, and Language Processing}, 29:3451--3460.

\bibitem[{Kahn et~al.(2020)Kahn, Rivi{\`{e}}re, Zheng, Kharitonov, Xu, Mazar{\'{e}}, Karadayi, Liptchinsky, Collobert, Fuegen, Likhomanenko, Synnaeve, Joulin, Mohamed, and Dupoux}]{DBLP:conf/icassp/KahnRZKXMKLCFLS20}
Jacob Kahn, Morgane Rivi{\`{e}}re, Weiyi Zheng, Evgeny Kharitonov, Qiantong Xu, Pierre{-}Emmanuel Mazar{\'{e}}, Julien Karadayi, Vitaliy Liptchinsky, Ronan Collobert, Christian Fuegen, Tatiana Likhomanenko, Gabriel Synnaeve, Armand Joulin, Abdelrahman Mohamed, and Emmanuel Dupoux. 2020.
\newblock \href {https://doi.org/10.1109/ICASSP40776.2020.9052942} {Libri-light: {A} benchmark for {ASR} with limited or no supervision}.
\newblock In \emph{2020 {IEEE} International Conference on Acoustics, Speech and Signal Processing, {ICASSP} 2020, Barcelona, Spain, May 4-8, 2020}, pages 7669--7673. {IEEE}.

\bibitem[{Kudo and Richardson(2018)}]{kudo-richardson-2018-sentencepiece}
Taku Kudo and John Richardson. 2018.
\newblock \href {https://doi.org/10.18653/v1/D18-2012} {{S}entence{P}iece: A simple and language independent subword tokenizer and detokenizer for neural text processing}.
\newblock In \emph{Proceedings of the 2018 Conference on Empirical Methods in Natural Language Processing: System Demonstrations}, pages 66--71, Brussels, Belgium. Association for Computational Linguistics.

\bibitem[{Lakhotia et~al.(2021)Lakhotia, Kharitonov, Hsu, Adi, Polyak, Bolte, Nguyen, Copet, Baevski, Mohamed, and Dupoux}]{lakhotia-etal-2021-generative}
Kushal Lakhotia, Eugene Kharitonov, Wei-Ning Hsu, Yossi Adi, Adam Polyak, Benjamin Bolte, Tu-Anh Nguyen, Jade Copet, Alexei Baevski, Abdelrahman Mohamed, and Emmanuel Dupoux. 2021.
\newblock \href {https://doi.org/10.1162/tacl_a_00430} {On generative spoken language modeling from raw audio}.
\newblock \emph{Transactions of the Association for Computational Linguistics}, 9:1336--1354.

\bibitem[{Le et~al.(2023)Le, Gong, Wang, Pino, Lecouteux, and Schwab}]{pmlr-v202-le23a}
Phuong-Hang Le, Hongyu Gong, Changhan Wang, Juan Pino, Benjamin Lecouteux, and Didier Schwab. 2023.
\newblock \href {https://proceedings.mlr.press/v202/le23a.html} {Pre-training for speech translation: {CTC} meets optimal transport}.
\newblock In \emph{Proceedings of the 40th International Conference on Machine Learning}, volume 202 of \emph{Proceedings of Machine Learning Research}, pages 18667--18685. PMLR.

\bibitem[{Liu et~al.(2020)Liu, Zhu, Zhang, and Zong}]{liu2020bridging}
Yuchen Liu, Junnan Zhu, Jiajun Zhang, and Chengqing Zong. 2020.
\newblock Bridging the modality gap for speech-to-text translation.
\newblock \emph{arXiv preprint arXiv:2010.14920}.

\bibitem[{Nguyen and Salazar(2019)}]{nguyen-salazar-2019-transformers}
Toan~Q. Nguyen and Julian Salazar. 2019.
\newblock \href {https://aclanthology.org/2019.iwslt-1.17} {Transformers without tears: Improving the normalization of self-attention}.
\newblock In \emph{Proceedings of the 16th International Conference on Spoken Language Translation}, Hong Kong. Association for Computational Linguistics.

\bibitem[{Ott et~al.(2019)Ott, Edunov, Baevski, Fan, Gross, Ng, Grangier, and Auli}]{ott-etal-2019-fairseq}
Myle Ott, Sergey Edunov, Alexei Baevski, Angela Fan, Sam Gross, Nathan Ng, David Grangier, and Michael Auli. 2019.
\newblock \href {https://doi.org/10.18653/v1/N19-4009} {fairseq: A fast, extensible toolkit for sequence modeling}.
\newblock In \emph{Proceedings of the 2019 Conference of the North {A}merican Chapter of the Association for Computational Linguistics (Demonstrations)}, pages 48--53, Minneapolis, Minnesota. Association for Computational Linguistics.

\bibitem[{Papi et~al.(2023)Papi, Gaido, Pilzer, and Negri}]{papi2023good}
Sara Papi, Marco Gaido, Andrea Pilzer, and Matteo Negri. 2023.
\newblock When good and reproducible results are a giant with feet of clay: The importance of software quality in nlp.
\newblock \emph{arXiv preprint arXiv:2303.16166}.

\bibitem[{Park et~al.(2019)Park, Chan, Zhang, Chiu, Zoph, Cubuk, and Le}]{park19e_interspeech}
Daniel~S. Park, William Chan, Yu~Zhang, Chung-Cheng Chiu, Barret Zoph, Ekin~D. Cubuk, and Quoc~V. Le. 2019.
\newblock \href {https://doi.org/10.21437/Interspeech.2019-2680} {{SpecAugment: A Simple Data Augmentation Method for Automatic Speech Recognition}}.
\newblock In \emph{Proc. Interspeech 2019}, pages 2613--2617.

\bibitem[{Pasad et~al.(2023)Pasad, Shi, and Livescu}]{pasad2023comparative}
Ankita Pasad, Bowen Shi, and Karen Livescu. 2023.
\newblock Comparative layer-wise analysis of self-supervised speech models.
\newblock In \emph{ICASSP 2023-2023 IEEE International Conference on Acoustics, Speech and Signal Processing (ICASSP)}, pages 1--5. IEEE.

\bibitem[{Peng et~al.(2023)Peng, Kim, Wu, Yan, Arora, Chen, Tang, Shon, Sridhar, and Watanabe}]{peng23b_interspeech}
Yifan Peng, Kwangyoun Kim, Felix Wu, Brian Yan, Siddhant Arora, William Chen, Jiyang Tang, Suwon Shon, Prashant Sridhar, and Shinji Watanabe. 2023.
\newblock \href {https://doi.org/10.21437/Interspeech.2023-1194} {{A Comparative Study on E-Branchformer vs Conformer in Speech Recognition, Translation, and Understanding Tasks}}.
\newblock In \emph{Proc. INTERSPEECH 2023}, pages 2208--2212.

\bibitem[{Polyak et~al.(2021)Polyak, Adi, Copet, Kharitonov, Lakhotia, Hsu, Mohamed, and Dupoux}]{polyak21_interspeech}
Adam Polyak, Yossi Adi, Jade Copet, Eugene Kharitonov, Kushal Lakhotia, Wei-Ning Hsu, Abdelrahman Mohamed, and Emmanuel Dupoux. 2021.
\newblock \href {https://doi.org/10.21437/Interspeech.2021-475} {{Speech Resynthesis from Discrete Disentangled Self-Supervised Representations}}.
\newblock In \emph{Proc. Interspeech 2021}, pages 3615--3619.

\bibitem[{Post(2018)}]{post-2018-call}
Matt Post. 2018.
\newblock \href {https://doi.org/10.18653/v1/W18-6319} {A call for clarity in reporting {BLEU} scores}.
\newblock In \emph{Proceedings of the Third Conference on Machine Translation: Research Papers}, pages 186--191, Brussels, Belgium. Association for Computational Linguistics.

\bibitem[{Radford et~al.(2023)Radford, Kim, Xu, Brockman, McLeavey, and Sutskever}]{radford2023robust}
Alec Radford, Jong~Wook Kim, Tao Xu, Greg Brockman, Christine McLeavey, and Ilya Sutskever. 2023.
\newblock Robust speech recognition via large-scale weak supervision.
\newblock In \emph{International Conference on Machine Learning}, pages 28492--28518. PMLR.

\bibitem[{Rei et~al.(2022)Rei, C.~de Souza, Alves, Zerva, Farinha, Glushkova, Lavie, Coheur, and Martins}]{rei-etal-2022-comet}
Ricardo Rei, Jos{\'e}~G. C.~de Souza, Duarte Alves, Chrysoula Zerva, Ana~C Farinha, Taisiya Glushkova, Alon Lavie, Luisa Coheur, and Andr{\'e} F.~T. Martins. 2022.
\newblock \href {https://aclanthology.org/2022.wmt-1.52} {{COMET}-22: Unbabel-{IST} 2022 submission for the metrics shared task}.
\newblock In \emph{Proceedings of the Seventh Conference on Machine Translation (WMT)}, pages 578--585, Abu Dhabi, United Arab Emirates (Hybrid). Association for Computational Linguistics.

\bibitem[{Sennrich et~al.(2016)Sennrich, Haddow, and Birch}]{sennrich-etal-2016-neural}
Rico Sennrich, Barry Haddow, and Alexandra Birch. 2016.
\newblock \href {https://doi.org/10.18653/v1/P16-1162} {Neural machine translation of rare words with subword units}.
\newblock In \emph{Proceedings of the 54th Annual Meeting of the Association for Computational Linguistics (Volume 1: Long Papers)}, pages 1715--1725, Berlin, Germany. Association for Computational Linguistics.

\bibitem[{Vaswani et~al.(2017)Vaswani, Shazeer, Parmar, Uszkoreit, Jones, Gomez, Kaiser, and Polosukhin}]{vaswani2017attention}
Ashish Vaswani, Noam Shazeer, Niki Parmar, Jakob Uszkoreit, Llion Jones, Aidan~N Gomez, {\L}ukasz Kaiser, and Illia Polosukhin. 2017.
\newblock Attention is all you need.
\newblock \emph{Advances in neural information processing systems}, 30.

\bibitem[{Wang et~al.(2020)Wang, Tang, Ma, Wu, Okhonko, and Pino}]{wang-etal-2020-fairseq}
Changhan Wang, Yun Tang, Xutai Ma, Anne Wu, Dmytro Okhonko, and Juan Pino. 2020.
\newblock \href {https://aclanthology.org/2020.aacl-demo.6} {Fairseq {S}2{T}: Fast speech-to-text modeling with fairseq}.
\newblock In \emph{Proceedings of the 1st Conference of the Asia-Pacific Chapter of the Association for Computational Linguistics and the 10th International Joint Conference on Natural Language Processing: System Demonstrations}, pages 33--39, Suzhou, China. Association for Computational Linguistics.

\bibitem[{Wang et~al.(2021)Wang, Wu, Gu, and Pino}]{wang21s_interspeech}
Changhan Wang, Anne Wu, Jiatao Gu, and Juan Pino. 2021.
\newblock \href {https://doi.org/10.21437/Interspeech.2021-2027} {{CoVoST 2 and Massively Multilingual Speech Translation}}.
\newblock In \emph{Proc. Interspeech 2021}, pages 2247--2251.

\bibitem[{Wu et~al.(2023)Wu, Kim, Watanabe, Han, McDonald, Weinberger, and Artzi}]{DBLP:conf/icassp/WuKWHMWA23}
Felix Wu, Kwangyoun Kim, Shinji Watanabe, Kyu~Jeong Han, Ryan McDonald, Kilian~Q. Weinberger, and Yoav Artzi. 2023.
\newblock \href {https://doi.org/10.1109/ICASSP49357.2023.10096988} {Wav2seq: Pre-training speech-to-text encoder-decoder models using pseudo languages}.
\newblock In \emph{{IEEE} International Conference on Acoustics, Speech and Signal Processing {ICASSP} 2023, Rhodes Island, Greece, June 4-10, 2023}, pages 1--5. {IEEE}.

\bibitem[{Yan et~al.(2024)Yan, Chang, Anastasopoulos, Fujita, and Watanabe}]{10447926}
Brian Yan, Xuankai Chang, Antonios Anastasopoulos, Yuya Fujita, and Shinji Watanabe. 2024.
\newblock \href {https://doi.org/10.1109/ICASSP48485.2024.10447926} {Cross-modal multi-tasking for speech-to-text translation via hard parameter sharing}.
\newblock In \emph{ICASSP 2024 - 2024 IEEE International Conference on Acoustics, Speech and Signal Processing (ICASSP)}, pages 11941--11945.

\bibitem[{Zeghidour et~al.(2021)Zeghidour, Luebs, Omran, Skoglund, and Tagliasacchi}]{zeghidour2021soundstream}
Neil Zeghidour, Alejandro Luebs, Ahmed Omran, Jan Skoglund, and Marco Tagliasacchi. 2021.
\newblock Soundstream: An end-to-end neural audio codec.
\newblock \emph{IEEE/ACM Transactions on Audio, Speech, and Language Processing}, 30:495--507.

\bibitem[{Zhang et~al.(2022{\natexlab{a}})Zhang, Haddow, and Sennrich}]{zhang2022revisiting}
Biao Zhang, Barry Haddow, and Rico Sennrich. 2022{\natexlab{a}}.
\newblock Revisiting end-to-end speech-to-text translation from scratch.
\newblock In \emph{International Conference on Machine Learning}, pages 26193--26205. PMLR.

\bibitem[{Zhang et~al.(2023{\natexlab{a}})Zhang, Haddow, and Sennrich}]{zhang-etal-2023-efficient}
Biao Zhang, Barry Haddow, and Rico Sennrich. 2023{\natexlab{a}}.
\newblock \href {https://doi.org/10.18653/v1/2023.eacl-main.166} {Efficient {CTC} regularization via coarse labels for end-to-end speech translation}.
\newblock In \emph{Proceedings of the 17th Conference of the European Chapter of the Association for Computational Linguistics}, pages 2264--2276, Dubrovnik, Croatia. Association for Computational Linguistics.

\bibitem[{Zhang et~al.(2023{\natexlab{b}})Zhang, Ye, Ko, Wang, and Zhou}]{zhang-etal-2023-dub}
Dong Zhang, Rong Ye, Tom Ko, Mingxuan Wang, and Yaqian Zhou. 2023{\natexlab{b}}.
\newblock \href {https://doi.org/10.18653/v1/2023.findings-acl.447} {{DUB}: Discrete unit back-translation for speech translation}.
\newblock In \emph{Findings of the Association for Computational Linguistics: ACL 2023}, pages 7147--7164, Toronto, Canada. Association for Computational Linguistics.

\bibitem[{Zhang et~al.(2022{\natexlab{b}})Zhang, Zhou, Ao, Liu, Dai, Li, and Wei}]{zhang-etal-2022-speechut}
Ziqiang Zhang, Long Zhou, Junyi Ao, Shujie Liu, Lirong Dai, Jinyu Li, and Furu Wei. 2022{\natexlab{b}}.
\newblock \href {https://doi.org/10.18653/v1/2022.emnlp-main.108} {{S}peech{UT}: Bridging speech and text with hidden-unit for encoder-decoder based speech-text pre-training}.
\newblock In \emph{Proceedings of the 2022 Conference on Empirical Methods in Natural Language Processing}, pages 1663--1676, Abu Dhabi, United Arab Emirates. Association for Computational Linguistics.

\end{thebibliography}

\end{document}